\documentclass[11pt]{article}

\usepackage[final]{acl}

\usepackage{times}
\usepackage{latexsym}
\usepackage[T1]{fontenc}
\usepackage[utf8]{inputenc}
\usepackage{microtype}
\usepackage{inconsolata}
\usepackage{graphicx}
\usepackage{booktabs}
\usepackage{amsmath}
\usepackage{array}
\usepackage{multirow}
\usepackage[table]{xcolor}
\newcommand{\hc}[1]{\cellcolor{blue!10}\textbf{#1}}
\usepackage{float}
\usepackage{fvextra}
\usepackage{xcolor}
\usepackage{hyperref}

\definecolor{darkblue}{rgb}{0.0, 0.0, 0.55}
\newif\iffigs
\figstrue
\graphicspath{{figs/}}

\newcommand{\rtr}{Rewrite-then-Reason}
\newcommand{\etr}{Extract-then-Reason}
\newcommand{\Pa}{\ensuremath{P_\alpha}}
\newcommand{\Pb}{\ensuremath{P_\beta}}
\newcommand{\Pg}{\ensuremath{P_\gamma}}
\newcommand{\dataset}{\textsc{NarrativeShield-SDoH}}

\title{Same Illness, Different Voice: Restating the Patient’s Account Narrows Narrative Anchoring in Clinical Language Models}

\author{
  \textbf{Prabhjot Singh} \\
  University of Texas at Austin \\
  Austin, Texas, USA \\
  \href{mailto:prabhjot.singh@utexas.edu}{\color{darkblue}\texttt{prabhjot.singh@utexas.edu}}
  \And
  \textbf{Pritam Deka} \\
  Queen's University Belfast \\
  Belfast, United Kingdom \\
  \href{mailto:pdeka01@qub.ac.uk}{\color{darkblue}\texttt{pdeka01@qub.ac.uk}}
  \And
  \textbf{Vijay Chennareddy} \\
  Middlesex University \\
  London, United Kingdom \\
  \href{mailto:Vc381@live.mdx.ac.uk}{\color{darkblue}\texttt{Vc381@live.mdx.ac.uk}}
}

\begin{document}
\maketitle

\begin{abstract}
Clinical language models are evaluated on vignettes written in a clinician's voice, but patients describe the same illness in their own. We pair 1,000 MedQA cases with three retellings that carry the same audited original facts, in a clinical, a socioeconomic and a cultural voice, and evaluate 12 open models (3B to 24B, including two medical fine-tunes). Every model is less accurate when the patient speaks in a marked voice: 26\% more errors, a pooled penalty of $-9.6$ accuracy points (95\% CI $[-11.0, -8.2]$), with no detectable reduction at larger scale. The penalty is not explained by any retelling property we measured (length, omitted age, MedQA split, added clinical content) and persists when the marked narrative contains the original vignette verbatim; because generation settings differ between voices, we read it as sensitivity to these marked retellings, not proof that register alone is the cause. Chain-of-thought and explicit debiasing instructions together remove only about a fifth of the gap. Having the same model first rewrite the patient's account as a clinical vignette, then answer from the rewrite, halves the gap while raising marked-voice accuracy by 3.9 points (means over the 11 models that can run the pipeline). A matched pipeline using a structured schema instead of prose closes the same gap but is 7.1 points less accurate in all 11 models, and adding reasoning agents or tools costs more still. Rewriting is not free: it recovers more facts than extraction but also invents plausible details, which we quantify. In our experiments, the representation a model reasons over matters more than how much it reasons.\footnote{Code, data and analysis scripts: \url{https://github.com/prabhjotschugh/narrativeshield-sdoh-medqa}}
\end{abstract}

\section{Introduction}
\label{sec:intro}

Language models are increasingly proposed for diagnostic decision support and are almost always evaluated on vignettes written by clinicians for clinicians \citep{jin2020disease,singhal2023large}. Patients do not talk like vignettes. They describe symptoms in the words they have, through the constraints they live with, and alongside the people who brought them in. If a model's answer depends on \emph{how} the facts are told rather than \emph{which} facts are told, its errors fall on the patients who sound least like a textbook.

Most work on clinical bias in language models manipulates explicit identity: stating race, gender or income and measuring whether recommendations change \citep{zack2024assessing,omiye2023large,pfohl2024toolbox,rawat2024diversitymedqa,poulain2024bias,omar2025sociodemographic}. Non-clinical surface content such as typos or informal language also shifts recommendations \citep{gourabathina2025medium}, and dialect alone elicits covert prejudice in general-purpose models \citep{hofmann2024ai}. Missing is a paired, fact-audited measurement of how a patient's own voice affects diagnostic accuracy across many models, a test of whether the effect survives the obvious confounds, and an account of what actually removes it.

We call this failure mode \textbf{narrative anchoring}: a change in a model's answer to the same clinical facts that tracks how the narrative is told; we operationalize it as the accuracy penalty on marked retellings relative to a clinical-voice retelling of the same case. Appendix~\ref{app:example} gives the full case in all three voices. A patient with frontotemporal dementia says, in a culturally framed voice, \emph{``I do not see phantoms or spirits that are not there''}; the clinical vignette says ``no hallucinations''. With the clinical voice, 8 of 11 models choose Pick's disease; with the cultural voice, 8 of 11 choose Lewy body dementia, a diagnosis defined by visual hallucinations. The metaphor is read as a symptom.

We build \dataset{}: 1,000 MedQA-USMLE vignettes, each retold in a clinical voice (\Pa), a socioeconomic voice (\Pb) and a cultural voice (\Pg) under an audited fact-preservation constraint, and evaluate 12 open models under four prompting conditions and five representation conditions. Our findings:

\begin{itemize}\setlength\itemsep{1pt}
\item \textbf{Anchoring is universal.} All 12 models are less accurate on both marked voices; every control-versus-marked McNemar test is significant after Bonferroni correction within each model (23 of 24 also across all 24 tests). Larger models are more accurate, but we detect no reduction in anchoring with size.
\item \textbf{Robust to the confounds we measured.} The gap does not vary with retelling length, MedQA split, or the amount of added clinical content under any of three judges, survives explicit-age and clean subsets, and appears even when the marked narrative contains the full original vignette. We cannot yet separate voice from generation settings (Limitations), so we describe this as sensitivity to marked retellings.
\item \textbf{Prompting barely helps.} Chain-of-thought raises accuracy for all voices without narrowing the gap; an explicit instruction to ignore the patient's voice removes 14\% of it, and combined with chain-of-thought 22\%.
\item \textbf{Representation does.} \rtr{} halves the gap and raises accuracy. A matched pipeline with a schema closes the same gap at a cost of 7.1 accuracy points; agents and tools lower accuracy further.
\end{itemize}

The model that rewrites is the same model that anchors. A model that gains little from an instruction to ignore a patient's voice can still restate that patient's account as a clinician would, and then reasons better. Our results are consistent with narrative anchoring being a problem of how models read patients, and with its remedy lying in what reaches the reasoner.

\section{Related Work}
\label{sec:related}

\paragraph{Demographic bias in clinical LLMs.} Explicit identity cues change model outputs in clinical vignettes \citep{zack2024assessing,omiye2023large}, benchmarks perturb stated demographics to measure it \citep{pfohl2024toolbox,rawat2024diversitymedqa}, larger or medically tuned models are not reliably less biased \citep{poulain2024bias}, and sociodemographic framing shifts care recommendations at scale \citep{omar2025sociodemographic}. Our voices state no race or income figure; the channel is how the story is told.

\paragraph{Non-clinical content and robustness.} Typos, colorful language and gender cues alter LLM triage recommendations \citep{gourabathina2025medium}; MedFuzz perturbs MedQA adversarially \citep{ness2024medfuzz}; irrelevant context degrades reasoning in general \citep{shi2023large}. We contribute fully paired retellings of whole cases, a confound audit, and a controlled comparison of mitigations.

\paragraph{Dialect and register.} Language technology encodes social hierarchies through how people speak \citep{blodgett2020language}: dialect predicts toxicity labels \citep{sap2019risk}, degrades NLU \citep{ziems2022value,ziems2023multivalue}, and elicits covert prejudice in LLM decisions \citep{hofmann2024ai}. We use \emph{register} in the sociolinguistic sense of situationally conditioned variety \citep{biber2009register}, extended to the non-clinical content patients naturally include.

\paragraph{Mitigation.} Instructions to avoid discrimination reduce some biases \citep{tamkin2023evaluating,ganguli2023capacity}, but chain-of-thought rationales can be steered by features they never mention \citep{turpin2023language}. System 2 Attention regenerates the context before answering \citep{weston2023system}, and LLMs are capable clinical extractors \citep{agrawal2022large}. We test both ideas, prose regeneration and structured extraction, in a controlled comparison with the evaluated model as its own rewriter.

\section{Benchmark}
\label{sec:benchmark}

\paragraph{Construction.} We filter MedQA-USMLE \citep{jin2020disease} for patient-vignette questions about diagnosis, treatment or investigation that contain patient-reported history, and draw 1,000 by stratified sampling over exam step and answer letter (Appendix~\ref{app:filters}). \texttt{gemini-3.1-flash-lite} retells each vignette in three voices in independent calls: clinical (\Pa, temperature 0.25), socioeconomic, framed by financial and occupational strain (\Pb, 0.90), and cultural, framed by family, tradition and somatic metaphor (\Pg, 0.90). Prompts forbid demographic labels and formulaic markers and rotate opening styles (Appendix~\ref{app:personas}). A separate auditor at temperature 0, without access to the generation prompt, checks that every original fact appears in the retelling; failures are regenerated.

\paragraph{What the voices change.} A preservation audit does not rule out additions or omissions, so we characterize the released retellings directly (Table~\ref{tab:properties}) and test each property in \S\ref{sec:voice}. Marked retellings are 2.2 to 2.6 times longer than \Pa{}; most add some clinically relevant detail (a home remedy, a self-medication, an exposure), though counts depend on the judge; first-person narrators often give the age in words or indirectly. The retellings therefore carry the same \emph{original} facts, audited, but are not minimal pairs. Throughout, ``voice'' denotes this whole bundle of differences between retellings (register, narrative person, length, added detail and generation temperature), not an isolated factor. In 116 questions the generator appended the original vignette verbatim after the narrative, which yields a natural superset condition.

\begin{table}[t]
\centering
\fontsize{10}{12.5}\selectfont
\setlength{\tabcolsep}{4pt}
\begin{tabular}{lccc}
\toprule
\cellcolor{blue!10} & \cellcolor{blue!10}$\boldsymbol{P_\alpha}$ & \cellcolor{blue!10}$\boldsymbol{P_\beta}$ & \cellcolor{blue!10}$\boldsymbol{P_\gamma}$ \\
\midrule
Length (words; original 137) & 117 & 261 & 301 \\
Explicit numeric age (\%) & 99.9 & 50.7 & 52.1 \\
$\geq$1 relevant addition (\%) & & & \\
\quad judge F & 22.6 & 85.9 & 79.3 \\
\quad judge P & 37.5 & 94.2 & 90.9 \\
\quad judge L & 61.1 & 91.2 & 82.3 \\
Original appended verbatim (\%) & 0.2 & 8.4 & 5.1 \\
\midrule
Facts preserved, clinicians (\%) & 100 & 100 & 100 \\
Realism, clinicians (1 to 5) & 4.89 & 4.64 & 4.19 \\
\bottomrule
\end{tabular}
\caption{Properties of the retellings. Judges F, P, L: \texttt{gemini-3.8-flash}, \texttt{gemini-3.1-pro-preview}, \texttt{gemini-3.1-flash-lite} (the generator's model), each listing content absent from the original and flagging items that could change diagnosis or management (Appendix~\ref{app:judges}).}
\label{tab:properties}
\end{table}

\paragraph{Human check.} Three physicians (an attending with 20 years of experience, a resident and an intern), independent of construction, rated 100 questions (300 retellings). No rater flagged an omitted or altered fact or a register mismatch in any of 900 ratings; these items have no variance, so we report them as a screening result rather than as reliability \citep{yanagita2024aigenerated}. Realism is ordered \Pa{} $>$ \Pb{} $>$ \Pg{} for every rater (Kruskal-Wallis $H=138.7$, $p<10^{-4}$), but agreement on the scale is low (Krippendorff's $\alpha$ \citep{krippendorff2004content} 0.226 excluding a rater who gave every item 5; $-0.014$ with all three), and falls from \Pa{} to \Pg{} \citep{kellert2026structured}. We therefore treat the marked voices as controlled \emph{register templates}, not as portraits of real communities.

\section{Experimental Setup}
\label{sec:setup}

\paragraph{Models.} Twelve open models in five families, all instruction-tuned checkpoints from the Hugging Face hub and referred to throughout by these exact names: Llama-3.2-3B-Instruct and Llama-3.1-8B-Instruct \citep{grattafiori2024llama3}; medgemma-4b-it \citep{sellergren2025medgemma}; Phi-4-mini-instruct \citep{abouelenin2025phi4mini}; Qwen2.5-7B-Instruct and Qwen2.5-14B-Instruct \citep{qwen2024qwen25}; Qwen3-8B with thinking disabled \citep{yang2025qwen3}; Mistral-7B-Instruct-v0.3 \citep{jiang2023mistral}; Ministral-8B-Instruct-2410 \citep{mistral2024ministral}; Mistral-Small-24B-Instruct-2501 \citep{mistral2025small}; gemma-3-12b-it \citep{gemmateam2025gemma3}; and the medical fine-tune Llama3-OpenBioLLM-8B \citep{pal2024openbiollm}. Decoding is greedy; every model answers all 3,000 retellings in every condition (compute is reported in the Compute paragraph of the Ethical Considerations).

\paragraph{Prompting conditions.} \textbf{B1} answers directly. \textbf{B2} reasons step by step first \citep{wei2022chain}. \textbf{B3} adds a system instruction to ignore how the patient describes symptoms, their health literacy and their cultural framing. \textbf{B4} combines B2 and B3.

\iffigs
\begin{figure*}[t]
\centering
\includegraphics[width=0.99\textwidth]{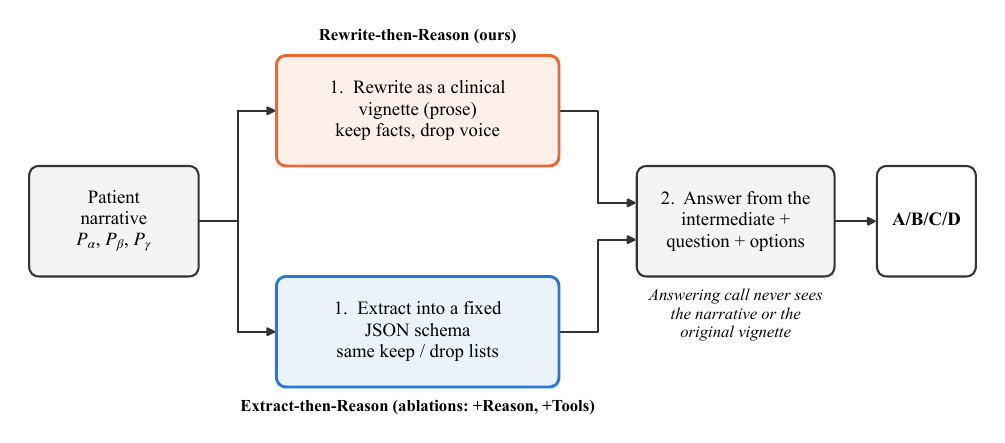}
\caption{Representation conditions. Both calls use the evaluated model; the answering call sees only the intermediate, the item's question sentence and the options. The two pipelines differ in the intermediate and in its keep/drop specification (Section~\ref{sec:setup}).}
\label{fig:method}
\end{figure*}
\fi

\paragraph{Representation conditions.} The evaluated model first converts the narrative into an intermediate, then answers from the intermediate, the item's question sentence and the options, without seeing the narrative (Figure~\ref{fig:method}; prompts in Appendix~\ref{app:prompts}). In \textbf{\rtr}, the intermediate is a third-person clinical vignette: the prompt asks to keep the age, sex, every symptom (present and denied) with its timing, every number, all history, medications and remedies taken, exposures and findings; to drop finances, work and family pressure, emotion and metaphor; and to add nothing and suggest no diagnosis. In \textbf{\etr}, the intermediate is a fixed JSON schema with a similar keep and drop list (it has no dedicated field for remedies taken or exposures, and its prompt strips home-remedy references; fields: age, sex, chief complaint, onset, symptoms present and absent, vitals, examination, labs, imaging, medications, history, allergies, one sentence of other context); if the JSON does not parse, the answering call receives the raw extraction text. The schema route is extended with a structured reasoning agent (\textbf{+Reason}) and, on top, a rule-based tool router (lab ranges, Wells, CURB-65, GCS, Apgar, a 47-drug knowledge base) with an answer-fallback agent (\textbf{+Tools}). \textbf{Story+Schema} answers in one call from the narrative plus the extracted schema.

\paragraph{Metrics.} Accuracy per voice ($n=1{,}000$). Our primary estimand is the \textbf{gap} $\mathrm{Acc}(\Pa)-\frac{1}{2}[\mathrm{Acc}(\Pb)+\mathrm{Acc}(\Pg)]$, which is unbiased and additive, so we use it for averages, regressions and comparisons between methods. We also report the \textbf{Narrative Anchoring Gap} $\mathrm{NAG}=\mathrm{Acc}(\Pa)-\min(\mathrm{Acc}(\Pb),\mathrm{Acc}(\Pg))$, which targets the worse-served voice; because a minimum of noisy estimates is biased upward, we use NAG only per model, with paired bootstrap intervals and a within-question permutation null (mean 0.005, 95th percentile 0.027). The \textbf{pooled effect} is the coefficient of a marked-voice indicator in a linear probability model with question-by-model fixed effects, clustered by question. Per-model comparisons use paired bootstrap intervals over questions (2,000 resamples); ``$k$/11'' counts models whose interval excludes zero. We also report Cohen's $\kappa$ between the letters chosen for \Pa{} and each marked voice, and transitions: the share of \emph{victim} items (\Pa{} correct, marked wrong under B1) a method recovers, and the share of marked items correct under B1 that it breaks. Llama3-OpenBioLLM-8B cannot produce the intermediates reliably (\S\ref{sec:costs}); we exclude it from mitigation averages and report 12-model averages, which lead to the same conclusions, in Appendix~\ref{app:sets}.

\section{Anchoring Is Universal}
\label{sec:pervasive}

\iffigs
\begin{figure}[b!]
\centering
\includegraphics[width=\columnwidth]{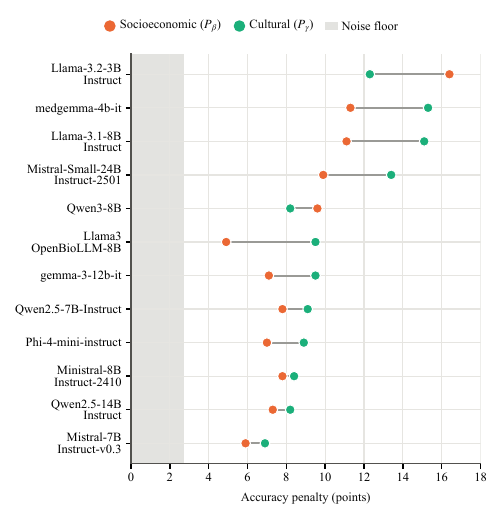}
\caption{Accuracy penalty of each marked voice versus the clinical voice under direct prompting, sorted by NAG. Every penalty exceeds the permutation-null band (shaded; 95th percentile 0.027).}
\label{fig:anchoring}
\end{figure}
\fi

Under direct prompting all 12 models are most accurate on the clinical voice (Figure~\ref{fig:anchoring}). Mean accuracy falls from 0.635 (\Pa) to 0.546 (\Pb) and 0.531 (\Pg), so the error rate rises from 36.5\% to 46.2\%: \textbf{26\% more wrong answers for the same audited original facts}. NAG ranges from 0.069 to 0.164, every interval excludes zero (Table~\ref{tab:app_nag_prompt}), and the smallest is 2.5 times the 95th percentile of the null. All 24 control-versus-marked McNemar tests are significant after Bonferroni correction over the three pairs of each model; over all 24 tests jointly ($\alpha=0.0021$), 23 remain significant, the exception being Llama3-OpenBioLLM-8B's \Pa{}-\Pb{} comparison ($p=0.003$; Appendix~\ref{app:mcnemar}). The pooled effect is $-9.6$ points $[-11.0, -8.2]$, and the cultural voice is the hardest in 10 of 12 models.

Model size predicts accuracy ($r=0.51$) but not NAG ($r=-0.06$, $p=0.85$; 12 models from 3B to 24B, so a modest scale effect could go undetected). The largest model, Mistral-Small-24B-Instruct-2501, has the fourth-largest NAG (0.134) and the medical medgemma-4b-it the second-largest (0.153). Models also contradict themselves: Cohen's $\kappa$ between the answers given to the same facts in different voices averages 0.57.

\section{Testing Alternative Explanations}
\label{sec:voice}

Marked retellings differ from \Pa{} in more than voice, so we test each measurable difference (Figure~\ref{fig:robust}).

\iffigs
\begin{figure}[H]
\centering
\includegraphics[width=0.5\textwidth]{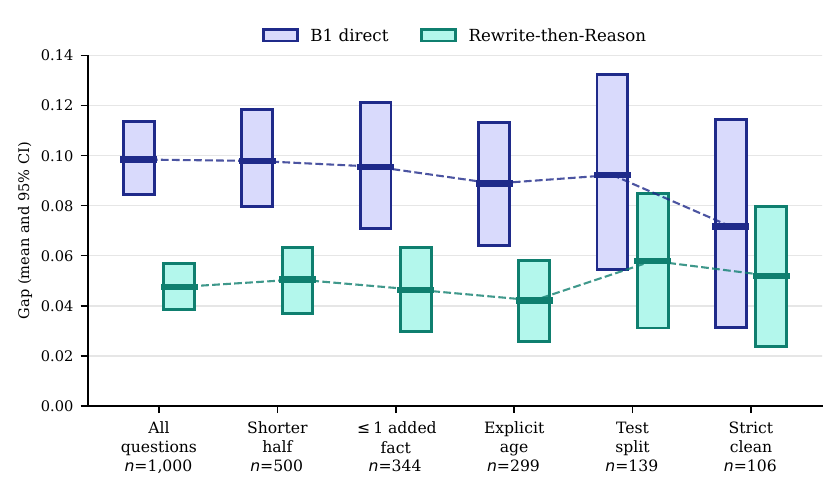}
\caption{Gap on controlled subsets under direct prompting and \rtr{} (11 models), ordered by subset size. Boxes span the 95\% CI, thick lines mark the mean, and dashed lines connect the means. Rewriting lowers the mean gap in every subset; the intervals overlap in the three smallest.}
\label{fig:robust}
\end{figure}
\fi

\textbf{Added content.} Under all three judges the gap is flat in the amount of clinically relevant added content: regression slopes of the per-question gap on the count are $-0.005$ ($p=0.57$, judge F), $-0.003$ ($p=0.63$, P) and $+0.007$ ($p=0.44$, L). Judge L shares a model with the generator; the two independent judges alone give the same result. Where neither marked retelling adds more than one relevant item (344 questions), the gap is 0.094, against 0.096 overall. The judges agree on which retellings add content (86.8\%, $\kappa=0.70$, and 78.9\%, $\kappa=0.51$, against judge F), although their counts differ.

\textbf{Length.} The gap is the same for the shorter and longer halves of retellings (0.094 and 0.098), and length has no effect in a joint regression with additions and age ($p=0.55$).

\textbf{Omitted age.} Age is the fact the marked voices most often leave implicit, and it explains a small part of the gap: each marked retelling without an explicit numeric age adds 2.1 points ($p=0.03$). With the age explicit in both marked retellings the gap is 0.085 (299 questions), 89\% of the full effect; on the strictest subset, with explicit age and at most one relevant addition per retelling (106 questions), it is 0.065 $[0.026, 0.105]$.

\textbf{The superset test.} In 84 \Pb{} retellings the original vignette was appended verbatim, so the marked input contains every original sentence plus the voice. The \Pa{}-versus-\Pb{} gap on these items is still 0.068 $[0.032, 0.106]$. Missing original information cannot explain this, since every original sentence is present; what remains is the added narrative, including its voice and any extra detail. These items are not a random sample, and \Pa{} was generated separately, so we read the test as suggestive evidence rather than an estimate of the full effect.

\textbf{MedQA split.} The gap is 0.090 on test-split items and 0.097 on training items (difference $-0.007$, CI $[-0.049, 0.034]$). The test split has been public since 2020 and has only 139 items, so this rules out only a large memorization effect.

\section{What Removes It}
\label{sec:mitigation}

\begin{table*}[t]
\centering
\fontsize{9.5}{12.5}\selectfont
\setlength{\tabcolsep}{3pt}
\begin{tabular}{>{\raggedright\arraybackslash}p{2.4cm}cccccccccc}
\toprule
\cellcolor{blue!10} &
\cellcolor{blue!10}$\boldsymbol{P_\alpha}$ &
\hc{Marked} &
\hc{Overall} &
\hc{Gap} &
\cellcolor{blue!10}\shortstack[c]{\textbf{Pooled effect}\\\textbf{[95\% CI]}} &
\cellcolor{blue!10}\shortstack[c]{\textbf{Gap}\\$\boldsymbol{\downarrow}$} &
\cellcolor{blue!10}\shortstack[c]{$\boldsymbol{\Delta}$\textbf{Acc}\\\textbf{vs B1}} &
\cellcolor{blue!10}\shortstack[c]{$\boldsymbol{\Delta}$\\\textbf{Marked}} &
\cellcolor{blue!10}$\boldsymbol{\kappa}$ &
\cellcolor{blue!10}\shortstack[c]{\textbf{Recovered}\\\textbf{/ Broken}} \\
\midrule
B1 direct & .645 & .547 & .579 & .098 & $-9.8$ [$-11.3$, $-8.4$] & & & & .57 & \\
B2 CoT & .676 & .586 & .616 & .090 & $-9.0$ [$-10.3$, $-7.7$] & 0/11 & $+3.7$ (8/0) & $+4.0$ & .55 & 40.9 / 17.4 \\
B3 debias & .632 & .547 & .576 & .085 & $-8.5$ [$-9.9$, $-7.1$] & 3/11 & $-0.4$ (2/2) & $+0.1$ & .59 & 18.1 / \phantom{0}8.6 \\
B4 CoT+debias & .663 & .587 & .612 & .077 & $-7.7$ [$-8.9$, $-6.5$] & 3/11 & $+3.3$ (8/1) & $+4.0$ & .56 & 44.2 / 18.4 \\
\midrule
Schema (\etr) & .564 & .515 & .531 & .049 & $-4.9$ [$-5.9$, $-3.9$] & 7/11 & $-4.9$ (0/10) & $-3.2$ & .64 & 43.4 / 27.5 \\
\quad +Reason & .554 & .513 & .527 & .041 & $-4.1$ [$-5.0$, $-3.1$] & 10/11 & $-5.3$ (0/8) & $-3.4$ & .58 & 43.8 / 31.2 \\
\quad +Tools & .535 & .495 & .508 & .040 & $-4.0$ [$-5.0$, $-3.1$] & 7/11 & $-7.1$ (0/11) & $-5.2$ & .58 & 41.7 / 34.2 \\
Story+Schema & .626 & .561 & .583 & .065 & $-6.5$ [$-7.7$, $-5.3$] & 7/11 & $+0.3$ (2/1) & $+1.4$ & .64 & 35.4 / 15.1 \\
\textbf{\rtr} & .633 & \textbf{.586} & \textbf{.602} & \textbf{.047} & $\mathbf{-4.7}$ [$-5.7$, $-3.8$] & \textbf{8/11} & $\mathbf{+2.2}$ (6/1) & $\mathbf{+3.9}$ & \textbf{.66} & \textbf{56.8} / 19.8 \\
\bottomrule
\end{tabular}
\caption{Main results, means over the 11 models that produce usable intermediates. Pooled effect: marked-voice penalty in accuracy points. Gap $\downarrow$: models with a significant gap reduction relative to B1. $\Delta$Acc: overall accuracy change in points (models significantly better / worse). Recovered: share of B1 victim items answered correctly; Broken: share of marked items correct under B1 that become wrong. Per-model values in Appendix~\ref{app:permodel}.}
\label{tab:main}
\end{table*}

\iffigs
\begin{figure}[h]
\centering
\includegraphics[width=\columnwidth]{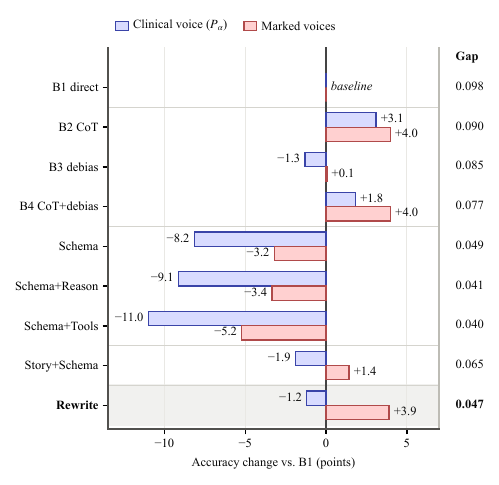}
\caption{Accuracy change relative to B1 for the clinical and marked voices (means over 11 models), with the resulting gap on the right. Prompting and \rtr{} raise marked-voice accuracy; every schema variant lowers accuracy for both voices, so its smaller gap comes from levelling down.}
\label{fig:methods}
\end{figure}
\fi

\subsection{Prompting barely helps}

Chain-of-thought raises accuracy for both the clinical voice (+3.1 points) and the marked voices (+4.0) but narrows the gap significantly in none of 11 models (Table~\ref{tab:main}, Figure~\ref{fig:methods}). An explicit instruction to ignore the patient's voice (B3) narrows the gap by 14\% (significant in 3 models), but only by levelling down: clinical-voice accuracy falls by 1.3 points while marked voices move by +0.1. Combining both (B4), the strongest prompt we could construct, removes 22\% of the gap and still leaves a penalty of $-7.7$ points. On diagnosis questions, B4 barely moves the gap (0.102 to 0.091). In our tests, a single instruction to disregard the patient's voice did little (we tried one wording; others may differ).

\subsection{Rewriting halves it and improves accuracy}

\rtr{} cuts the pooled marked-voice penalty from $-9.8$ to $-4.7$ points. The gap shrinks significantly in 8 of 11 models and grows in none. Unlike every other gap-reducing method here, it levels up: marked voices gain 3.9 points (\Pb{} +4.2, \Pg{} +3.6; significant in 8 of 11), the clinical voice moves by $-1.2$, and overall accuracy rises by 2.2 points (better in 6 models, worse only in Ministral-8B-Instruct-2410). It recovers 56.8\% of victim items, the most of any condition, and gives the most consistent answers across voices ($\kappa$ 0.57 to 0.66). On diagnosis questions it halves the gap (0.102 to 0.054). The improvement is largest where anchoring is worst: per-model gap reduction correlates with the direct-prompting gap at $r=0.86$ (Llama-3.1-8B-Instruct 0.131 to 0.037; Llama-3.2-3B-Instruct 0.143 to 0.053), and is unrelated to size ($r=-0.01$); Figure~\ref{fig:heatmap} and Table~\ref{tab:app_rewrite} (Appendix) show every model and method.

Against B4, \rtr{} reaches the same marked-voice accuracy (0.586 vs.\ 0.587) with a 39\% smaller gap (0.047 vs.\ 0.077; smaller in 4 models, larger in none), at 1.0 point lower overall accuracy, because chain-of-thought additionally lifts the clinical voice. Against Story+Schema, it has a smaller gap and higher accuracy (+1.9 points, better in 5 models, worse in none); against B3 it has a smaller gap (0.047 vs.\ 0.085) and 2.6 points higher accuracy.

\subsection{Prose versus schema: a controlled comparison}

\iffigs
\begin{figure*}[t]
\centering
\includegraphics[width=\textwidth]{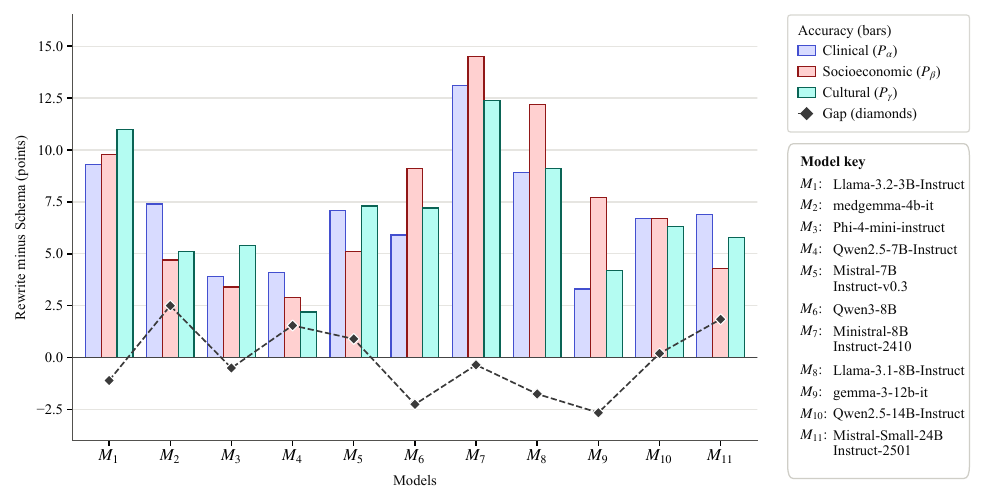}
\caption{Same model, same two calls, same question sentence; the intermediate (and its similar but not identical keep/drop specification) differs. Bars show accuracy of \rtr{} minus \etr{} for each voice; the diamond shows the difference in gap. The rewrite is more accurate on every voice in all 11 models, while the gap difference stays within $\pm 3$ points of zero.}
\label{fig:prose}
\end{figure*}
\fi

\etr{} and \rtr{} use the same model, calls and question sentence and differ in the intermediate (a JSON schema versus prose; the keep/drop specifications are similar but not identical, Section~\ref{sec:setup}). Both close the same gap (0.049 vs.\ 0.047; no model significant either way), but the rewrite is \textbf{7.1 points more accurate overall and 7.1 points more accurate on the marked voices, in all 11 of 11 models} (Figure~\ref{fig:prose}). This is not a parse-failure artefact: 95.9\% of extractions parse, and on the questions where all three parse the rewrite is still 6.8 points more accurate, in 11 of 11 models (Appendix~\ref{app:pipeline}). Both representations reduce the gap equally; the schema, as specified here, also discards more clinical information (including remedies and exposures, which the rewrite is told to keep), so the accuracy difference reflects this schema and prompt rather than structure in general. A judge audit confirms this: the schema retains 0.80, 0.73 and 0.72 of the narrative's facts for \Pa, \Pb{} and \Pg{}, the rewrite 0.96, 0.84 and 0.81. Every schema variant levels down, leaving marked voices 3.2 to 5.2 points \emph{worse} than direct prompting; its gap closes because the clinical voice falls further.

\subsection{More machinery is not better}

Adding a reasoning agent to the schema pipeline changes the gap by 0.008, inside the noise, and costs 0.4 points. Adding the tool router and fallback lowers overall accuracy from 0.579 (B1) and 0.531 (schema) to 0.508, worse than B1 in all 11 models, and breaks 34.2\% of previously correct marked answers, while its gap (0.040) is close to the rewrite's (0.047), which achieves it at far higher accuracy (0.602). The tools fire on about 47\% of questions, and on those items the pipeline is 3.3 points less accurate than without them. The multi-agent design is thus dominated by a two-call rewrite.

\subsection{Controls for the mitigation}
\label{sec:controls}

\textbf{Question sentence.} The answering call receives the item's question sentence, whereas under direct prompting the model reads the patient's version of it (only 30\% of \Pb{} and 58\% of \Pg{} narratives contain it verbatim). On the 207 items where both marked narratives already contain the original sentence, \rtr{} still reduces the gap by 3.3 points $[0.9, 5.7]$ and raises marked accuracy by 3.3 $[0.9, 5.4]$; elsewhere the reduction is 5.6 points. Part of the gain therefore comes from the canonical question, and most does not. The prose-versus-schema comparison is unaffected, since both pipelines receive the same sentence. \textbf{Split.} On test-split items the gap reduction is 3.4 points $[0.3, 6.4]$ (5.4 on training items), and the advantage over the schema is 7.1 points on both. \textbf{Subsets.} The rewrite gap stays between 0.03 and 0.06 on every controlled subset (Figure~\ref{fig:robust}).

\section{What the Rewrite Costs, and What Remains}
\label{sec:costs}

\textbf{Rewrites fill gaps.} Rewriting compresses the marked narratives from 261 and 301 words to 174 and 167, close to the clinical voice, and recovers more facts than extraction, but it also adds what a clinical vignette ``should'' contain. Among marked rewrites, 34\% state a numeric age that the narrative does not give (median 8 years off), 29\% (\Pb) and 19\% (\Pg) contain a vital sign, lab value or dose that is not in the narrative (8\% for \Pa), and templated negatives (``no known drug allergies'') are common. Twelve percent of invented values match the original MedQA vignette, so models occasionally recall the textbook case. Invention rates range from 5\% (gemma-3-12b-it) to 39\% (Ministral-8B-Instruct-2410) of marked rewrites. Additions are not harmless: in the audited sample, each clinically relevant added item lowers the probability of a correct answer by 1.0 point ($p=0.01$, controlling for question, model and recall). They therefore cannot explain the rewrite's gains, which hold on test-split items, but they are a real cost.

\textbf{What remains.} The rewrite NAG exceeds the null's 95th percentile in all 11 instruction-following models (0.037 to 0.088; Table~\ref{tab:app_nag_repr}). The residual is largely accounted for by lost facts: fact recall of the rewrite is 0.96 for \Pa{} but 0.84 and 0.81 for the marked voices, and in the audited sample the within-question marked-voice penalty under \rtr{} ($-2.0$ points) vanishes once per-item recall is controlled ($+0.7$, $p=0.68$; recall coefficient 0.21, $p=0.006$). What is left of narrative anchoring after rewriting is a reading problem: marked narratives are harder to restate completely.

\textbf{A failure case.} Llama3-OpenBioLLM-8B anchors like every other model under direct prompting (NAG 0.095) but cannot produce usable intermediates: 23\% of its rewrite answers and 45\% of its schema answers parse, and its mean accuracy across voices falls to 0.10 and 0.20 (Table~\ref{tab:app_acc_repr}). For this model, fine-tuning appears to have removed the instruction following the intervention requires; the other medical model, medgemma-4b-it, benefits from rewriting like the general models. This single collapse is also why the 12-model means in Appendix~\ref{app:sets} differ in sign from the 11-model means: rewriting raises mean overall accuracy by 2.2 points over the 11 models that can run it, but OpenBioLLM loses 37.9 points, which turns the 12-model mean change into about $-1.2$ points (Table~\ref{tab:app_sets}). The gap results are unaffected.

\section{Discussion}
\label{sec:discussion}

\paragraph{Reading, not knowing.} B3 and B4 suggest that models gain little from an instruction to ignore a patient's voice. \rtr{} shows that the same models can restate that voice as a clinician would, and that doing so helps; what remains after rewriting tracks the facts the rewrite failed to carry over. The superset result suggests that the added narrative interferes even when every original clinical sentence is present. Together, these results are consistent with narrative anchoring being a failure of how models read patients rather than of what they know, which would explain why changing the representation works where changing the instruction does not; isolating voice from generation settings is needed to confirm this.

\paragraph{For evaluation.} Single-presentation benchmarks cannot detect this failure. Paired retellings make it measurable, and reporting the gap beside accuracy exposes methods that close it by making the favored group worse, as the debiasing prompt and every schema pipeline here do.

\paragraph{For systems.} A clinical intake step that restates patient-authored text in clinical prose, performed by the same model, is cheap and model-agnostic. Because rewrites also invent plausible details, such a step should keep the original narrative visible to the clinician and should be constrained, for instance to state only values present in the source. Combining rewriting with chain-of-thought, which lifts the clinical voice that rewriting does not, is a natural next step.

\section{Conclusion}

Retelling the same cases in a patient's marked voice lowers the accuracy of 12 open clinical language models by about ten points, in every model and with no detectable reduction at larger scale. The effect survives controls for length, omitted age, added content and MedQA split, and persists when the full clinical vignette is present, though we cannot yet separate voice from generation settings. Prompting barely moves it. Restating the patient's account as a clinical vignette before answering halves it and improves accuracy, while a matched pipeline built on a structured schema loses information and accuracy, and more agents and tools only cost more. Fair clinical language models need a better interface between a patient's words and the model's reasoning.

\section*{Limitations}

\textbf{Generation confound.} \Pa{} was generated at temperature 0.25 and the marked voices at 0.90 by the same model. Our controls address the measurable consequences (length, added content, omitted age, the superset condition), but we did not generate a register-neutral paraphrase of each case at the same temperature (0.90) and a matched length, which would isolate the voice from generation noise. The clinical voice is also third-person prose while the marked voices are first-person, so narrative person varies with voice. That control requires generating and clinician-validating a new persona set, which we consider outside the scope of this paper and leave to future work. Until it is run, our claim is that the penalty is not explained by any property we measured, not that register or voice is its only cause; we therefore describe our findings as sensitivity to these marked retellings. The superset test covers 84 non-random items and shows a smaller gap (0.068) than the full effect, so it bounds rather than isolates the voice effect.

\textbf{Synthetic voices.} The retellings come from one generator and were validated by clinicians for fact preservation, not by sociolinguists or community members for authenticity, and realism agreement is low. We measure sensitivity to controlled register templates, not how any real community speaks.

\textbf{Judges.} The added-content audit uses three LLM judges (one sharing the generator's model); the schema and rewrite audits use a single judge (\texttt{gemini-3.8-flash}). No audit was checked against clinicians.

\textbf{Mitigation.} \rtr{} halves rather than removes the gap, invents details in a substantial share of marked rewrites, slightly lowers clinical-voice accuracy, depends on instruction following, and part of its gain over direct prompting comes from the canonical question sentence (\S\ref{sec:controls}). The schema and rewrite keep/drop specifications are similar but not identical, so the prose-versus-schema difference reflects this schema and prompt, not structure in general. It is a research finding about representations, not a validated clinical tool.

\textbf{Scope.} We evaluate open models up to 24B parameters, the class hospitals can deploy on premises, not proprietary frontier models, on single-turn multiple-choice questions with greedy decoding. The MedQA test split is public, so our split analysis cannot rule out small memorization effects.

\section*{Ethical Considerations}

\dataset{} derives from MedQA-USMLE, a public, de-identified corpus of examination questions; all narratives are synthetic and no patient data are used. Three physicians volunteered as annotators, rated only synthetic text, and asked that any compensation be directed to further research; no IRB review was required. Synthetic renderings of socioeconomic or cultural voice risk caricature; we forbade formulaic markers and report realism ratings openly, without claiming representativeness. The finding that voice shifts diagnoses could be misused to craft narratives that steer a deployed model; we judge that documenting the vulnerability, with an analysis of what mitigates it, is more protective. Rewriting patient text with a language model can introduce plausible but false details, as we show; such rewrites must not replace the patient's own words in a clinical record. Neither the models nor the pipelines evaluated are intended for unsupervised clinical use. The dataset is released for non-commercial research with attribution to MedQA. Generative AI assisted with language editing; all claims, numbers and citations were checked against code and data.
\paragraph{Compute.} Each run (model $\times$ condition) is a single-GPU job on an NVIDIA GH200 node (TACC Vista) that writes results per question and resumes after interruption. Chain-of-thought conditions allow 1,280 new tokens, direct conditions 128, rewrites 512. Total inference was about 1,113 GPU-hours (282 jobs, including failed and re-run jobs). LLM-judge audits (about 17,000 calls) need no GPU. The anonymous repository (footnote~1) contains the 3,000 retellings, all prompts, per-question result files for every model and condition, and the analysis scripts for every table and figure.

\section*{Acknowledgements}

We express our sincere gratitude to the medical professionals who generously volunteered their clinical expertise to annotate and validate the dataset for this work:
\textbf{Dr.~Roopam Deka}, DM (Assistant Professor, Department of Pathology, All India Institute of Medical Sciences, Guwahati);
\textbf{Dr.~Benjina Ahmed}, MBBS (Junior Resident, Department of Pathology and Lab Medicine, All India Institute of Medical Sciences, Guwahati); and
\textbf{Dr.~Meghna Kashyap}, MBBS (Intern, Laxmi Chandravansi Medical College and Hospital).
Their rigorous evaluation of clinical fact preservation and narrative fidelity was fundamental to ensuring the benchmark's quality.

\clearpage
\bibliography{custom}

\clearpage
\appendix

\section{Filtering and Sampling}
\label{app:filters}

We apply six filters as one boolean mask over the pooled MedQA-USMLE train and test splits (seed 42): (1) a patient-vignette stem matching the regular expression below; (2) at least 300 characters; (3) at least one of 49 clinical signal terms; (4) none of 21 mechanism, pathophysiology, anatomy, genetics or histology phrases; (5) at least one of 24 diagnosis, treatment or investigation phrases; (6) at least one of 27 narrative-richness markers (e.g., ``she states'', ``for the past''). Of 2,921 eligible questions we draw 1,000 by proportional stratified sampling over exam step and answer letter. Matching item text to the public splits gives 861 training and 139 test items. Full term lists are in the released code.

\begin{quote}
\normalsize
\begin{verbatim}
^(A|An)\s+\d+[\-\s]?
(year|month|week|day)[\-\s]?old
\end{verbatim}
\end{quote}

\section{Retelling Prompts and Audit}
\label{app:personas}

Each retelling is generated in its own call. \Pa{} keeps clinical register and asks for the original length ($\pm$10\%); \Pb{} and \Pg{} prompts asked for 10 to 25\% more words, but the generator wrote longer narratives (Table~\ref{tab:properties}), and we report measured lengths. Marked prompts forbid generic markers (e.g., ``traditional herbal teas'', ``my family insisted'') and assign one of five opening styles per attempt. The auditor (temperature 0) extracts every fact from the original, checks its presence in any form, and returns pass or fail; all 3,000 released retellings passed.

\section{Worked Example}
\label{app:example}

Counts are the answers of the 11 instruction-following models (Table~\ref{tab:app_example}). All three keep the findings, including the denial of hallucinations. Under direct prompting, 8 of 11 models are correct in the clinical voice but only 2 of 11 in either marked voice, which most often yield Lewy body dementia, a diagnosis defined by hallucinations. The socioeconomic retelling even ends with the full original vignette, and models still fail on it. Restating the account as a clinical vignette first restores the answer for the cultural voice (8 of 11) and partly for the socioeconomic voice (5 of 11).

\paragraph{\Pa{} (clinical).}
\begin{quote}\small
A 53-year-old male shopkeeper presents for evaluation of an 8-month history of progressive behavioral changes and cognitive decline. He reports increasing irritability, executive dysfunction manifesting as financial mismanagement, and social disinhibition characterized by verbal aggression toward clientele. He exhibits paranoid ideation, specifically accusing a customer of witchcraft. Concurrently, he reports progressive anomia and word-finding difficulties. He denies recent head trauma, febrile illness, \textbf{hallucinations}, or involuntary movements. Past medical history is significant for type 2 diabetes mellitus, currently well-controlled. Family history is non-contributory. He denies tobacco or illicit substance use. Vitals are stable: BP 134/76 mm Hg, HR 88/min, and temperature 37.0$^{\circ}$C (98.6$^{\circ}$F). Physical examination reveals deficits in naming and executive planning. MMSE score is 26/30. Cranial nerve, motor, and sensory examinations are within normal limits, with symmetric tone and reflexes. What is the most likely diagnosis?
\end{quote}

\paragraph{\Pb{} (socioeconomic).}
\begin{quote}\small
\setlength{\parskip}{3pt}
If I can't run the shop, we're going to lose the place and everything we've got, and my son's already tired of me acting like this. My boy dragged me in here today because he's scared, and honestly, I'm scared too. I've been acting real nasty to the folks coming in to buy stuff lately---even accused one guy of putting a hex on my store, which I know sounds crazy, but I just haven't been right. I've been real short-tempered for about eight months now, and my head's just not working. I keep forgetting where I put the money, and the register is a mess because I can't keep the numbers straight. Words just don't come to me, and I'll be staring at a broom or a can of soup and just can't pull the name out of my head. I didn't come in sooner because every time I go to a doctor, that copay wipes out my whole week's profit, and I've just been chugging coffee and trying to push through. I don't have no fever, haven't taken any spills or hit my head, and \textbf{I ain't seeing things} or shaking in my limbs. I've got that sugar-diabetes, but I've been keeping that under control. I don't touch cigarettes or any of that bad stuff.

The nurse checked me over---said my blood pressure is 134/76, heart's beating 88 times a minute, and my temp is 98.6, so I'm fine there. She did that test where they ask you to name things and plan stuff out, and I really struggled with it. She said I got a 26 out of 30 on that brain test, whatever that means. She checked my nerves, said they looked fine, and my arms and legs are working right---no weakness, no stiff muscles, and my reflexes and feeling are all normal. What's going on with me?

A 53-year-old man is brought to the clinic by his son for the evaluation of unusual behavior. He is a shopkeeper by profession and sometimes behaves very rudely to the customers. Recently, he accused one of the customers of using black magic over his shop. He has been increasingly irritable, forgetting things, and having problems managing his finances over the past 8 months. He is also having difficulty finding words and recalling the names of objects during the conversation. There is no history of recent head trauma, fever, \textbf{hallucinations}, or abnormal limb movements. Past medical history is significant for a well-controlled type 2 diabetes mellitus. Family history is unremarkable. He does not smoke or use illicit drugs. Vital signs are stable with a blood pressure of 134/76 mm Hg, a heart rate of 88/min, and a temperature of 37.0$^{\circ}$C (98.6$^{\circ}$F). On physical examination, he has problems naming objects and planning tasks. Mini-mental state examination (MMSE) score is 26/30. Cranial nerve examination is normal. Muscle strength is normal in all 4 limbs with normal muscle tone and deep tendon reflexes. Sensory examination is also normal. What is the most likely diagnosis?
\end{quote}

\paragraph{\Pg{} (cultural).}
\begin{quote}\small
\setlength{\parskip}{3pt}
My duty has always been to serve those who enter my shop with respect and to ensure my ledger remains in perfect balance, yet these obligations have become heavy burdens I can no longer lift. My eldest son, who watches my accounts with a concerned eye, brought me here today because he saw the composure of my character slipping away; he told me, ``Father, the kindness that defines our name is being replaced by a shadow,'' and he insisted we seek counsel. For the past eight months, my mind has been failing its duty to the truth. I have spoken with unnecessary sharpness to my patrons, and once, I felt such confusion that I accused a guest of casting a dark enchantment to sour my shop's prosperity.

My inner state is fractured: the names of common things hide from my tongue like shy birds in a thicket, and my thoughts, once orderly, now behave as if the threads of a loom have been tangled by an unruly hand. Managing the coins and the books has become a labor where the numbers refuse to hold their place. My son and I have fasted and offered prayers for clarity, yet the fog remains. I have no fever to burn away this cloud, and my head carries no scars from any recent blow. \textbf{I do not see phantoms or spirits that are not there}, and my limbs remain quiet, without the involuntary dancing or shaking that would betray a weakness of the blood. My back is silent and without complaint, and my chest feels no pressure, nor does my breath struggle for air. My heart beats with a steady rhythm, not a panicked drum.

At fifty-three years of age, I have lived with the sugar in my blood for some time, which I manage with care, and I touch neither smoke nor foreign poisons. My pulse remains at 88 beats each minute, my pressure at 134/76 mm Hg, and my body heat is a neutral 37.0$^{\circ}$C (98.6$^{\circ}$F). Though I stand before you, my ability to name the objects of the world is frayed, and the map for my daily tasks is erased from my reach. The doctor says my inner focus, when measured on the score of thirty, is at twenty-six. My nerves of the face and the strength in my arms and legs hold their vigor and tone, and the signals from my skin are clear and true. Given these conditions, what is the most likely diagnosis?
\end{quote}

\paragraph{Options and answers.}
A.~Creutzfeldt--Jakob disease \quad B.~Huntington's disease \quad C.~Lewy body dementia \quad D.~Pick's disease. \quad \textbf{Gold:} D. Pick's disease (frontotemporal dementia).

\begin{table}[t]
\centering
\fontsize{9.5}{12.5}\selectfont
\setlength{\tabcolsep}{4pt}
\begin{tabular}{l>{\raggedright\arraybackslash}p{2.8cm}>{\raggedright\arraybackslash}p{2.8cm}}
\toprule
\cellcolor{blue!10} & \hc{Direct prompting (B1)} & \hc{\rtr} \\
\midrule
\Pa{} & Pick's 8, Lewy body 3 & Pick's 9, Lewy body 2 \\
\Pb{} & Lewy body 8, Pick's 2, Huntington's 1 & Lewy body 6, Pick's 5 \\
\Pg{} & Lewy body 8, Pick's 2, Huntington's 1 & Pick's 8, Lewy body 2, Creutzfeldt--Jakob 1 \\
\bottomrule
\end{tabular}
\caption{Answers of the 11 instruction-following models to MQ\_0153 (gold: Pick's disease) under direct prompting and \rtr{}, by voice. Entries give the number of models choosing each diagnosis.}
\label{tab:app_example}
\end{table}

\section{Added-Content Audit and Judge Agreement}
\label{app:judges}

For every question and voice, a judge receives the original vignette and the retelling, lists each piece of information in the retelling that the original does not contain or imply (ignoring wording and tone), and marks it clinically relevant if it could change diagnosis or management. Three judges audited all 3,000 items at temperature 0: \texttt{gemini-3.8-flash} (F, primary) and \texttt{gemini-3.1-pro-preview} (P), which did not generate the retellings, and \texttt{gemini-3.1-flash-lite} (L), the generator's model, included as a third view; we use several judges because single LLM judges have known biases \citep{zheng2023judging}. On ``at least one relevant addition'', agreement with judge F is 86.8\% ($\kappa=0.70$; within-voice $\kappa$ 0.50 to 0.58) for P and 78.9\% ($\kappa=0.51$) for L; Spearman correlations of counts are 0.76 and 0.58. P flags more additions in marked retellings and L more in \Pa{}; all three agree on the ordering \Pa{} $\ll$ \Pg{} $<$ \Pb{} and give a flat gap, as do F and P alone. Under the strict judge P only 8 questions have no relevant addition in either marked retelling, so subset analyses use judge F and regressions use all three.

\section{Full Results}
\label{app:permodel}

\iffigs
\begin{figure*}[t]
\centering
\includegraphics[width=\textwidth]{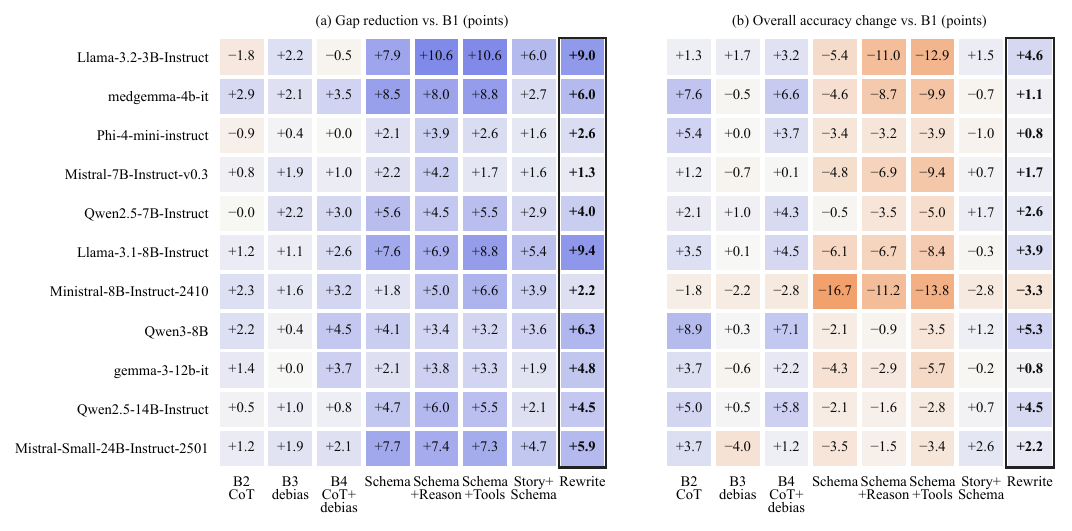}
\caption{Per-model change relative to B1 for every method (rows: models, ordered by size). (a) Gap reduction in points; blue is a smaller gap. (b) Change in overall accuracy in points; orange is a loss. The schema variants lose accuracy in every model, whereas \rtr{} gains or holds accuracy in all but Ministral-8B-Instruct-2410.}
\label{fig:heatmap}
\end{figure*}
\fi

Figure~\ref{fig:heatmap} gives the per-model change relative to B1 for every method. Table~\ref{tab:app_rewrite} compares \rtr{} with direct prompting model by model; Tables~\ref{tab:app_acc_prompt} and~\ref{tab:app_acc_repr} give accuracy per voice under the prompting and representation conditions; Tables~\ref{tab:app_nag_prompt} and~\ref{tab:app_nag_repr} give the corresponding NAG with 95\% intervals. Llama3-OpenBioLLM-8B is shown in every table but excluded from the mitigation averages in the main text, and bold NAG entries mark intervals that exclude zero.

\section{Other Model Sets}
\label{app:sets}

The main text averages over the 11 instruction-following models; Table~\ref{tab:app_sets} repeats the condition means with Llama3-OpenBioLLM-8B included (12 models), which lowers every method's overall accuracy; the conclusions about the gap are unchanged. The sign change for rewriting (overall accuracy $+2.2$ over 11 models, $-1.2$ over 12) is due to Llama3-OpenBioLLM-8B, which cannot produce usable intermediates and loses 37.9 points under rewriting (Section~\ref{sec:costs}); this reflects a failure of intermediate generation in one fine-tuned model, not of the representation idea.

\section{Significance Tests}
\label{app:mcnemar}
Table~\ref{tab:app_mcnemar} gives the McNemar test for each pair of voices under direct prompting. All significance statements in the paper are generated by the released analysis scripts. Across all 24 control-versus-marked tests jointly ($\alpha=0.05/24$), 23 are significant.

\section{Intermediate Audits}
\label{app:pipeline}

\textbf{Schema extraction.} For 100 questions per model and voice (3,600 audits), \texttt{gemini-3.8-flash} compared the extracted JSON with its narrative. When the JSON parses, it retains 0.75 of the narrative's facts on average (0.67 to 0.80 by model); 1.9\% of extractions lose nothing. Losses are mostly contextual: exposure details, negative findings, imaging modality. Parse rates are similar across voices (96.5\%, 96.0\% and 95.2\% for \Pa, \Pb, \Pg; Table~\ref{tab:app_parse}).

\textbf{Rewrites.} For a fixed sample of 100 questions, all three voices and all 12 models (3,600 audits), the same judge listed clinical facts of the narrative missing from the rewrite (non-clinical content the rewrite is told to drop does not count) and content in the rewrite that the narrative does not state or imply. Fact recall is 0.96 (\Pa), 0.84 (\Pb) and 0.81 (\Pg). Clinically relevant added items average 0.7, 3.1 and 2.5 per rewrite; about 30\% of those in marked rewrites are templated negatives and 12 to 20\% a stated age or sex. Deterministic checks over all 33,000 rewrites of the 11 instruction-following models give the age and measured-value rates in \S\ref{sec:costs}.

\raggedbottom
\section{Prompts}
\label{app:prompts}

Verbatim prompts for the answering conditions (B1--B4), \rtr{} and \etr{}; text in braces is filled in by the code. All other prompts (retelling generation, the fact-preservation auditor and judges, and the +Reason, +Tools and Story+Schema agents) are in the \texttt{prompts/} directory of the anonymous repository (footnote~1). The retellings were produced with a hosted API (\texttt{gemini-3.1-flash-lite}) whose models change over time, so we release the retellings themselves; a regeneration will not be bit-for-bit identical.

\paragraph{B1--B4.} The narrative is the full retelling, including the question sentence.
\begin{Verbatim}[breaklines=true,breakanywhere=true,fontsize=\scriptsize]
B1 user turn (no system prompt):
{narrative}

Options:
A: {option A}
B: {option B}
C: {option C}
D: {option D}

Which of the following is the correct option? Provide only the single correct letter.

B2 user turn (no system prompt):
{narrative}

Options:
A: {option A}
...
D: {option D}

Think step by step. Consider each option carefully based only on the clinical facts presented. Then state your final answer as: Answer: [letter]

B3 system turn:
Ignore how the patient describes their symptoms. Focus exclusively on the objective clinical facts, test results, and medical history. Do not let the patient's language style, health literacy, or cultural framing influence your reasoning.
B3 user turn: identical to the B1 user turn.

B4 system turn: identical to the B3 system turn.
B4 user turn: identical to the B2 user turn.
\end{Verbatim}

\paragraph{\rtr{}: rewrite call and answering call.}
\begin{Verbatim}[breaklines=true,breakanywhere=true,fontsize=\scriptsize]
Rewrite call, system turn:
You are a clinician writing up a patient's own account as a standard clinical case vignette.

Rewrite the patient narrative as a concise, third-person clinical vignette in standard medical terminology, the way a USMLE case is written.

KEEP, exactly and completely:
- the patient's age and sex (if the age is given only approximately, state it as given)
- every symptom, with its onset, duration, timing and course, including symptoms the patient denies
- every number: vitals, lab values, doses, durations, measurements
- all past medical, surgical, family, medication, allergy, exposure and social history that bears on health (e.g. smoking, alcohol, travel, occupation exposures, substances taken, remedies or medicines the patient took)
- all physical examination, laboratory and imaging findings

DROP: financial worries, insurance, work or family pressures, who brought the patient in, emotions, cultural or metaphorical phrasing, and any other wording that is about how the patient speaks rather than what happened clinically.

Do not add, infer or interpret anything that is not in the narrative. Do not suggest a diagnosis. Output only the vignette.

Rewrite call, user turn (maximum 512 new tokens):
PATIENT NARRATIVE:
{narrative}

Write the clinical vignette.

Answering call, system turn (maximum 16 new tokens):
Select the best answer letter (A, B, C, or D). Output exactly one letter.

Answering call, user turn:
CLINICAL VIGNETTE:
{rewrite}

QUESTION:
{question sentence}

OPTIONS:
  A: {option A}
  B: {option B}
  C: {option C}
  D: {option D}

Answer letter:
\end{Verbatim}

\paragraph{\etr{}: schema extraction.} The extraction call returns a JSON object with the fields listed in Section~\ref{sec:setup} (the full template is in the repository). The answering call is the \rtr{} answering call with ``CLINICAL FACTS: \{extracted JSON\}'' in place of the vignette.
\begin{Verbatim}[breaklines=true,breakanywhere=true,fontsize=\scriptsize]
Extraction call, system turn (maximum 600 new tokens):
You are a clinical information extraction system. Your only function is to extract objective clinical facts from a patient narrative and return them as a JSON object.

STRIP completely -- do not mention anywhere in output:
- How the patient phrases or describes their experience (lay vs. clinical vocabulary)
- Economic barriers: cannot afford, missed work, waited due to cost, no insurance
- Care-access delays caused by economics or geography
- Cultural or traditional medicine references: herbal tea, home remedy, family medicine
- Family-prompted care-seeking: mother insisted, husband brought me
- Emotional language: scared, worried, suffering, distressed
- Metaphorical or somatic framing: burning heat in belly, fire in my body, heavy spirit
- Any language signaling socioeconomic or cultural identity

PRESERVE exactly:
- Every number (age, gestational weeks, vitals, lab values, durations in days/hours, doses)
- Every symptom reported (present and absent/denied)
- All medications by pharmacological name (if patient uses a cultural name like "breath-guard" for an inhaler, record as "inhaler [type if inferrable]" -- do NOT preserve the cultural name)
- All past medical history, surgical history, allergies
- All physical exam findings and lab results

Return ONLY the JSON object (fields as in Section 4) -- no text before or after. Use null for missing scalars and [] for missing lists.

Extraction call, user turn:
PATIENT NARRATIVE:
{narrative}

Extract the clinical facts as JSON.
\end{Verbatim}

\begin{table*}[t]
\centering
\fontsize{9.5}{12.5}\selectfont
\setlength{\tabcolsep}{2pt}
\resizebox{\ifdim\width>\textwidth\textwidth\else\width\fi}{!}{%
\begin{tabular}{>{\raggedright\arraybackslash}p{2.5cm}cccccccc}
\toprule
\cellcolor{blue!10} & \multicolumn{2}{c}{\hc{B1 direct}} & \multicolumn{2}{c}{\hc{Rewrite-then-Reason}} & \hc{Gap reduction} & \cellcolor{blue!10}$\boldsymbol{\Delta}$ \textbf{overall acc.} & \cellcolor{blue!10}$\boldsymbol{\Delta}$ \textbf{marked} & \cellcolor{blue!10}$\boldsymbol{\kappa}$ \\
\cmidrule(lr){2-3}\cmidrule(lr){4-5}
\hc{Model} & \cellcolor{blue!10}$\boldsymbol{P_\alpha}$ \textbf{/ marked} & \hc{Gap} & \cellcolor{blue!10}$\boldsymbol{P_\alpha}$ \textbf{/ marked} & \hc{Gap} & \hc{(lower 95\% bound)} & \hc{[95\% CI]} & \hc{acc.} & \cellcolor{blue!10}\textbf{B1} $\to$ \textbf{Rewrite} \\
\midrule
Llama-3.2-3B-\allowbreak Instruct & .668 / .524 & .143 & .654 / .601 & .053 & .090 (.058) & +4.6 [+2.6, +6.6] & +7.6 & .546 $\to$ .609 \\
medgemma-4b-it & .667 / .534 & .133 & .638 / .565 & .073 & .060 (.029) & +1.1 [$-$0.8, +2.9] & +3.1 & .529 $\to$ .632 \\
Phi-4-mini-instruct & .562 / .482 & .080 & .553 / .499 & .053 & .026 ($-$.009) & +0.8 [$-$1.2, +2.9] & +1.7 & .529 $\to$ .536 \\
Qwen2.5-7B-\allowbreak Instruct & .614 / .530 & .084 & .613 / .569 & .044 & .041 (.013) & +2.6 [+0.9, +4.3] & +4.0 & .570 $\to$ .671 \\
Mistral-7B-\allowbreak Instruct-v0.3 & .530 / .466 & .064 & .538 / .487 & .051 & .013 ($-$.020) & +1.7 [$-$0.2, +3.6] & +2.1 & .532 $\to$ .578 \\
Qwen3-8B & .650 / .561 & .089 & .661 / .635 & .025 & .064 (.030) & +5.3 [+3.5, +7.1] & +7.4 & .553 $\to$ .688 \\
Ministral-8B-\allowbreak Instruct-2410 & .595 / .514 & .081 & .547 / .488 & .059 & .022 ($-$.009) & $-$3.3 [$-$5.4, $-$1.2] & $-$2.6 & .573 $\to$ .654 \\
Llama-3.1-8B-\allowbreak Instruct & .729 / .598 & .131 & .706 / .669 & .037 & .093 (.064) & +3.9 [+2.1, +5.7] & +7.0 & .566 $\to$ .700 \\
Llama3-OpenBioLLM-8B$^\dagger$ & .522 / .450 & .072 & .109 / .088 & .021 & .051 (.016) & $-$37.9 [$-$40.3, $-$35.4] & $-$36.2 & .476 $\to$ .333 \\
gemma-3-12b-it & .657 / .574 & .083 & .633 / .598 & .035 & .048 (.021) & +0.8 [$-$0.9, +2.6] & +2.4 & .572 $\to$ .704 \\
Qwen2.5-14B-\allowbreak Instruct & .672 / .595 & .077 & .687 / .655 & .032 & .045 (.019) & +4.5 [+2.8, +6.3] & +6.0 & .617 $\to$ .757 \\
Mistral-Small-24B-\allowbreak Instruct-2501 & .752 / .635 & .117 & .735 / .677 & .058 & .059 (.029) & +2.2 [+0.4, +4.0] & +4.2 & .636 $\to$ .737 \\
\bottomrule
\end{tabular}%
}
\caption{Rewrite-then-Reason against direct prompting, per model. A positive lower bound means the gap reduction is significant. Accuracy changes are in points. $\kappa$: agreement between the answers given for $P_\alpha$ and each marked voice. $^\dagger$OpenBioLLM cannot reliably produce the structured or rewritten intermediate and is excluded from mitigation averages.}
\label{tab:app_rewrite}
\end{table*}

\begin{table*}[t]
\centering
\fontsize{9}{12.5}\selectfont
\setlength{\tabcolsep}{6pt}
\resizebox{\ifdim\width>\textwidth\textwidth\else\width\fi}{!}{%
\begin{tabular}{>{\raggedright\arraybackslash}p{3.3cm}ccc|ccc|ccc|ccc}
\toprule
\cellcolor{blue!10} & \multicolumn{3}{c|}{\hc{B1 direct}} & \multicolumn{3}{c|}{\hc{B2 CoT}} & \multicolumn{3}{c|}{\hc{B3 debias}} & \multicolumn{3}{c}{\hc{B4 CoT+debias}} \\
\hc{Model} & \cellcolor{blue!10}$\boldsymbol{P_\alpha}$ & \cellcolor{blue!10}$\boldsymbol{P_\beta}$ & \cellcolor{blue!10}$\boldsymbol{P_\gamma}$ & \cellcolor{blue!10}$\boldsymbol{P_\alpha}$ & \cellcolor{blue!10}$\boldsymbol{P_\beta}$ & \cellcolor{blue!10}$\boldsymbol{P_\gamma}$ & \cellcolor{blue!10}$\boldsymbol{P_\alpha}$ & \cellcolor{blue!10}$\boldsymbol{P_\beta}$ & \cellcolor{blue!10}$\boldsymbol{P_\gamma}$ & \cellcolor{blue!10}$\boldsymbol{P_\alpha}$ & \cellcolor{blue!10}$\boldsymbol{P_\beta}$ & \cellcolor{blue!10}$\boldsymbol{P_\gamma}$ \\
\midrule
Llama-3.2-3B-\allowbreak Instruct & .668 & .504 & .545 & .693 & .513 & .549 & .670 & .552 & .546 & .703 & .552 & .557 \\
medgemma-4b-it & .667 & .554 & .514 & .724 & .651 & .589 & .648 & .539 & .533 & .710 & .632 & .591 \\
Phi-4-mini-instruct & .562 & .492 & .473 & .622 & .537 & .530 & .560 & .494 & .474 & .599 & .545 & .494 \\
Qwen2.5-7B-\allowbreak Instruct & .614 & .536 & .523 & .635 & .559 & .542 & .609 & .551 & .542 & .637 & .602 & .564 \\
Mistral-7B-\allowbreak Instruct-v0.3 & .530 & .471 & .461 & .537 & .492 & .469 & .511 & .479 & .452 & .525 & .475 & .466 \\
Qwen3-8B & .650 & .554 & .568 & .724 & .670 & .645 & .650 & .566 & .565 & .691 & .653 & .641 \\
Ministral-8B-\allowbreak Instruct-2410 & .595 & .517 & .511 & .562 & .515 & .492 & .562 & .495 & .500 & .546 & .507 & .487 \\
Llama-3.1-8B-\allowbreak Instruct & .729 & .618 & .578 & .756 & .640 & .633 & .723 & .617 & .588 & .757 & .654 & .650 \\
Llama3-OpenBioLLM-8B$^\dagger$ & .522 & .473 & .427 & .325 & .287 & .323 & .508 & .462 & .419 & .381 & .316 & .349 \\
gemma-3-12b-it & .657 & .586 & .562 & .684 & .625 & .606 & .651 & .578 & .559 & .654 & .622 & .595 \\
Qwen2.5-14B-\allowbreak Instruct & .672 & .599 & .590 & .719 & .656 & .637 & .670 & .605 & .601 & .725 & .666 & .645 \\
Mistral-Small-24B-\allowbreak Instruct-2501 & .752 & .653 & .618 & .781 & .691 & .662 & .699 & .628 & .575 & .750 & .669 & .640 \\
\bottomrule
\end{tabular}%
}
\caption{Accuracy per voice under the prompting conditions ($n=1{,}000$ per cell). $^\dagger$OpenBioLLM cannot reliably produce the structured or rewritten intermediate and is excluded from mitigation averages.}
\label{tab:app_acc_prompt}
\end{table*}

\begin{table*}[t]
\centering
\fontsize{9}{12.5}\selectfont
\setlength{\tabcolsep}{5pt}
\resizebox{\ifdim\width>\textwidth\textwidth\else\width\fi}{!}{%
\begin{tabular}{>{\raggedright\arraybackslash}p{3.3cm}ccc|ccc|ccc|ccc|ccc}
\toprule
\cellcolor{blue!10} & \multicolumn{3}{c|}{\hc{Schema}} & \multicolumn{3}{c|}{\hc{+Reason}} & \multicolumn{3}{c|}{\hc{+Tools}} & \multicolumn{3}{c|}{\hc{Story+Schema}} & \multicolumn{3}{c}{\hc{Rewrite}} \\
\hc{Model} & \cellcolor{blue!10}$\boldsymbol{P_\alpha}$ & \cellcolor{blue!10}$\boldsymbol{P_\beta}$ & \cellcolor{blue!10}$\boldsymbol{P_\gamma}$ & \cellcolor{blue!10}$\boldsymbol{P_\alpha}$ & \cellcolor{blue!10}$\boldsymbol{P_\beta}$ & \cellcolor{blue!10}$\boldsymbol{P_\gamma}$ & \cellcolor{blue!10}$\boldsymbol{P_\alpha}$ & \cellcolor{blue!10}$\boldsymbol{P_\beta}$ & \cellcolor{blue!10}$\boldsymbol{P_\gamma}$ & \cellcolor{blue!10}$\boldsymbol{P_\alpha}$ & \cellcolor{blue!10}$\boldsymbol{P_\beta}$ & \cellcolor{blue!10}$\boldsymbol{P_\gamma}$ & \cellcolor{blue!10}$\boldsymbol{P_\alpha}$ & \cellcolor{blue!10}$\boldsymbol{P_\beta}$ & \cellcolor{blue!10}$\boldsymbol{P_\gamma}$ \\
\midrule
Llama-3.2-3B-\allowbreak Instruct & .561 & .511 & .483 & .487 & .464 & .435 & .469 & .444 & .418 & .643 & .579 & .540 & .654 & .609 & .593 \\
medgemma-4b-it & .564 & .533 & .499 & .527 & .482 & .466 & .509 & .483 & .445 & .642 & .547 & .524 & .638 & .580 & .550 \\
Phi-4-mini-instruct & .514 & .470 & .441 & .504 & .481 & .446 & .506 & .462 & .443 & .542 & .488 & .468 & .553 & .504 & .495 \\
Qwen2.5-7B-\allowbreak Instruct & .572 & .550 & .537 & .549 & .503 & .517 & .527 & .500 & .496 & .611 & .565 & .547 & .613 & .579 & .559 \\
Mistral-7B-\allowbreak Instruct-v0.3 & .467 & .436 & .414 & .433 & .408 & .413 & .425 & .382 & .373 & .527 & .491 & .466 & .538 & .487 & .487 \\
Qwen3-8B & .602 & .558 & .550 & .619 & .576 & .551 & .594 & .544 & .530 & .638 & .590 & .580 & .661 & .649 & .622 \\
Ministral-8B-\allowbreak Instruct-2410 & .416 & .361 & .346 & .449 & .415 & .422 & .413 & .393 & .403 & .541 & .505 & .494 & .547 & .506 & .470 \\
Llama-3.1-8B-\allowbreak Instruct & .617 & .565 & .559 & .616 & .557 & .551 & .587 & .538 & .549 & .690 & .633 & .592 & .706 & .687 & .650 \\
Llama3-OpenBioLLM-8B$^\dagger$ & .186 & .200 & .204 & .386 & .381 & .353 & .375 & .353 & .340 & .061 & .096 & .078 & .109 & .087 & .089 \\
gemma-3-12b-it & .600 & .539 & .538 & .603 & .556 & .559 & .578 & .531 & .525 & .642 & .590 & .567 & .633 & .616 & .580 \\
Qwen2.5-14B-\allowbreak Instruct & .620 & .592 & .587 & .616 & .599 & .599 & .608 & .586 & .584 & .665 & .619 & .599 & .687 & .659 & .650 \\
Mistral-Small-24B-\allowbreak Instruct-2501 & .666 & .629 & .624 & .688 & .642 & .648 & .669 & .621 & .630 & .747 & .683 & .672 & .735 & .672 & .682 \\
\bottomrule
\end{tabular}%
}
\caption{Accuracy per voice under the representation conditions. Schema: Extract-then-Reason; +Reason and +Tools: its extensions; Story+Schema: narrative plus extracted schema; Rewrite: Rewrite-then-Reason. $^\dagger$OpenBioLLM cannot reliably produce the structured or rewritten intermediate and is excluded from mitigation averages.}
\label{tab:app_acc_repr}
\end{table*}

\begin{table*}[t]
\centering
\fontsize{9}{12.5}\selectfont
\setlength{\tabcolsep}{9pt}
\begin{tabular}{>{\raggedright\arraybackslash}p{3.6cm}cccc}
\toprule
\hc{Model} & \hc{B1 direct} & \hc{B2 CoT} & \hc{B3 debias} & \hc{B4 CoT+debias} \\
\midrule
Llama-3.2-3B-\allowbreak Instruct & \textbf{.164 [.133, .196]} & \textbf{.180 [.146, .214]} & \textbf{.124 [.099, .156]} & \textbf{.151 [.124, .185]} \\
medgemma-4b-it & \textbf{.153 [.123, .186]} & \textbf{.135 [.102, .167]} & \textbf{.115 [.090, .147]} & \textbf{.119 [.087, .151]} \\
Phi-4-mini-instruct & \textbf{.089 [.061, .121]} & \textbf{.092 [.064, .129]} & \textbf{.086 [.057, .117]} & \textbf{.105 [.072, .140]} \\
Qwen2.5-7B-\allowbreak Instruct & \textbf{.091 [.064, .122]} & \textbf{.093 [.063, .123]} & \textbf{.067 [.039, .097]} & \textbf{.073 [.042, .104]} \\
Mistral-7B-\allowbreak Instruct-v0.3 & \textbf{.069 [.042, .101]} & \textbf{.068 [.039, .101]} & \textbf{.059 [.030, .089]} & \textbf{.059 [.033, .091]} \\
Qwen3-8B & \textbf{.096 [.068, .130]} & \textbf{.079 [.049, .111]} & \textbf{.085 [.064, .119]} & \textbf{.050 [.024, .083]} \\
Ministral-8B-\allowbreak Instruct-2410 & \textbf{.084 [.061, .115]} & \textbf{.070 [.037, .105]} & \textbf{.067 [.044, .097]} & \textbf{.059 [.028, .094]} \\
Llama-3.1-8B-\allowbreak Instruct & \textbf{.151 [.120, .183]} & \textbf{.123 [.099, .156]} & \textbf{.135 [.105, .166]} & \textbf{.107 [.083, .142]} \\
Llama3-OpenBioLLM-8B$^\dagger$ & \textbf{.095 [.062, .127]} & \textbf{.038 [.003, .075]} & \textbf{.089 [.059, .123]} & \textbf{.065 [.028, .102]} \\
gemma-3-12b-it & \textbf{.095 [.065, .127]} & \textbf{.078 [.050, .108]} & \textbf{.092 [.065, .122]} & \textbf{.059 [.030, .086]} \\
Qwen2.5-14B-\allowbreak Instruct & \textbf{.082 [.059, .110]} & \textbf{.082 [.055, .108]} & \textbf{.069 [.046, .099]} & \textbf{.080 [.053, .109]} \\
Mistral-Small-24B-\allowbreak Instruct-2501 & \textbf{.134 [.106, .163]} & \textbf{.119 [.092, .148]} & \textbf{.124 [.095, .153]} & \textbf{.110 [.083, .138]} \\
\bottomrule
\end{tabular}
\caption{NAG $=\mathrm{Acc}(P_\alpha)-\min(\mathrm{Acc}(P_\beta),\mathrm{Acc}(P_\gamma))$ with 95\% paired bootstrap CIs (2,000 resamples of questions); bold: CI excludes zero. The permutation null has mean 0.005 and 95th percentile 0.027. Prompting conditions.}
\label{tab:app_nag_prompt}
\end{table*}
\begin{table*}[t]
\centering
\fontsize{9}{12.5}\selectfont
\setlength{\tabcolsep}{4pt}
\resizebox{\ifdim\width>\textwidth\textwidth\else\width\fi}{!}{%
\begin{tabular}{>{\raggedright\arraybackslash}p{3.0cm}ccccc}
\toprule
\hc{Model} & \hc{Schema} & \hc{+Reason} & \hc{+Tools} & \hc{Story+Schema} & \hc{Rewrite} \\
\midrule
Llama-3.2-3B-\allowbreak Instruct & \textbf{.078 [.051, .108]} & \textbf{.052 [.021, .082]} & \textbf{.051 [.021, .084]} & \textbf{.103 [.075, .132]} & \textbf{.061 [.035, .092]} \\
medgemma-4b-it & \textbf{.065 [.036, .094]} & \textbf{.061 [.034, .091]} & \textbf{.064 [.033, .094]} & \textbf{.118 [.090, .149]} & \textbf{.088 [.061, .115]} \\
Phi-4-mini-instruct & \textbf{.073 [.044, .102]} & \textbf{.058 [.027, .088]} & \textbf{.063 [.033, .095]} & \textbf{.074 [.046, .102]} & \textbf{.058 [.032, .090]} \\
Qwen2.5-7B-\allowbreak Instruct & \textbf{.035 [.010, .064]} & \textbf{.046 [.020, .079]} & \textbf{.031 [.009, .063]} & \textbf{.064 [.039, .090]} & \textbf{.054 [.030, .080]} \\
Mistral-7B-\allowbreak Instruct-v0.3 & \textbf{.053 [.027, .080]} & \textbf{.025 [.003, .055]} & \textbf{.052 [.030, .081]} & \textbf{.061 [.033, .089]} & \textbf{.051 [.031, .083]} \\
Qwen3-8B & \textbf{.052 [.028, .081]} & \textbf{.068 [.040, .097]} & \textbf{.064 [.037, .093]} & \textbf{.058 [.035, .088]} & \textbf{.039 [.012, .066]} \\
Ministral-8B-\allowbreak Instruct-2410 & \textbf{.070 [.047, .097]} & \textbf{.034 [.011, .062]} & .020 [$-$.002, .047] & \textbf{.047 [.024, .074]} & \textbf{.077 [.051, .104]} \\
Llama-3.1-8B-\allowbreak Instruct & \textbf{.058 [.036, .087]} & \textbf{.065 [.041, .097]} & \textbf{.049 [.025, .076]} & \textbf{.098 [.069, .126]} & \textbf{.056 [.030, .082]} \\
Llama3-OpenBioLLM-8B$^\dagger$ & $-$.014 [$-$.036, .017] & \textbf{.033 [.004, .064]} & \textbf{.035 [.008, .068]} & $-$.017 [$-$.037, .003] & \textbf{.022 [.004, .048]} \\
gemma-3-12b-it & \textbf{.062 [.042, .092]} & \textbf{.047 [.026, .079]} & \textbf{.053 [.031, .083]} & \textbf{.075 [.047, .106]} & \textbf{.053 [.029, .078]} \\
Qwen2.5-14B-\allowbreak Instruct & \textbf{.033 [.011, .062]} & .017 [$-$.003, .048] & \textbf{.024 [.004, .053]} & \textbf{.066 [.041, .091]} & \textbf{.037 [.017, .061]} \\
Mistral-Small-24B-\allowbreak Instruct-2501 & \textbf{.042 [.021, .068]} & \textbf{.046 [.025, .074]} & \textbf{.048 [.023, .077]} & \textbf{.075 [.053, .098]} & \textbf{.063 [.043, .087]} \\
\bottomrule
\end{tabular}%
}
\caption{NAG $=\mathrm{Acc}(P_\alpha)-\min(\mathrm{Acc}(P_\beta),\mathrm{Acc}(P_\gamma))$ with 95\% paired bootstrap CIs (2,000 resamples of questions); bold: CI excludes zero. The permutation null has mean 0.005 and 95th percentile 0.027. Representation conditions.}
\label{tab:app_nag_repr}
\end{table*}

\begin{table*}[t]
\centering
\fontsize{9.5}{12.5}\selectfont
\setlength{\tabcolsep}{8pt}
\resizebox{\ifdim\width>\textwidth\textwidth\else\width\fi}{!}{%
\begin{tabular}{>{\raggedright\arraybackslash}p{2.6cm}cccccccc}
\toprule
\cellcolor{blue!10} & \multicolumn{4}{c}{\hc{11 models (main text)}} & \multicolumn{4}{c}{\hc{12 models (with OpenBioLLM)}} \\
\cmidrule(lr){2-5}\cmidrule(lr){6-9}
\cellcolor{blue!10} & \cellcolor{blue!10}$\boldsymbol{P_\alpha}$ & \hc{Marked} & \hc{Overall} & \hc{Gap} & \cellcolor{blue!10}$\boldsymbol{P_\alpha}$ & \hc{Marked} & \hc{Overall} & \hc{Gap} \\
\midrule
B1 direct & .645 & .547 & .579 & .098 & .635 & .539 & .571 & .096 \\
B2 CoT & .676 & .586 & .616 & .090 & .647 & .563 & .591 & .084 \\
B3 debias & .632 & .547 & .576 & .085 & .622 & .538 & .566 & .083 \\
B4 CoT+debias & .663 & .587 & .612 & .077 & .640 & .566 & .590 & .074 \\
Schema & .564 & .515 & .531 & .049 & .532 & .489 & .503 & .043 \\
+Reason & .554 & .513 & .527 & .041 & .540 & .501 & .514 & .039 \\
+Tools & .535 & .495 & .508 & .040 & .522 & .482 & .495 & .039 \\
Story+Schema & .626 & .561 & .583 & .065 & .579 & .521 & .541 & .058 \\
Rewrite & .633 & .586 & .602 & .047 & .590 & .544 & .559 & .045 \\
\bottomrule
\end{tabular}%
}
\caption{Condition means on the 11 instruction-following models (left, as in the main table) and on all 12 models including OpenBioLLM (right). With OpenBioLLM included the rewrite still has the smallest gap among methods that do not lower marked-voice accuracy; its overall accuracy falls below B1 only because OpenBioLLM collapses under rewriting ($-37.9$ points).}
\label{tab:app_sets}
\end{table*}

\begin{table*}[t]
\centering
\fontsize{9.5}{12.5}\selectfont
\setlength{\tabcolsep}{14pt}
\begin{tabular}{lccc}
\toprule
\hc{Model} & \cellcolor{blue!10}$\boldsymbol{p(\alpha,\beta)}$ & \cellcolor{blue!10}$\boldsymbol{p(\alpha,\gamma)}$ & \cellcolor{blue!10}$\boldsymbol{p(\beta,\gamma)}$ \\
\midrule
Llama-3.2-3B-Instruct & $<$0.001$^*$ & $<$0.001$^*$ & 0.008$^*$ \\
medgemma-4b-it & $<$0.001$^*$ & $<$0.001$^*$ & 0.016$^*$ \\
Phi-4-mini-instruct & $<$0.001$^*$ & $<$0.001$^*$ & 0.236 \\
Qwen2.5-7B-Instruct & $<$0.001$^*$ & $<$0.001$^*$ & 0.409 \\
Mistral-7B-Instruct-v0.3 & $<$0.001$^*$ & $<$0.001$^*$ & 0.548 \\
Qwen3-8B & $<$0.001$^*$ & $<$0.001$^*$ & 0.379 \\
Ministral-8B-Instruct-2410 & $<$0.001$^*$ & $<$0.001$^*$ & 0.736 \\
Llama-3.1-8B-Instruct & $<$0.001$^*$ & $<$0.001$^*$ & 0.009$^*$ \\
Llama3-OpenBioLLM-8B$^\dagger$ & 0.003$^*$ & $<$0.001$^*$ & 0.003$^*$ \\
gemma-3-12b-it & $<$0.001$^*$ & $<$0.001$^*$ & 0.133 \\
Qwen2.5-14B-Instruct & $<$0.001$^*$ & $<$0.001$^*$ & 0.576 \\
Mistral-Small-24B-Instruct-2501 & $<$0.001$^*$ & $<$0.001$^*$ & 0.018 \\
\bottomrule
\end{tabular}
\caption{McNemar tests under B1. $^*$significant after Bonferroni correction over the three pairs per model ($\alpha=0.0167$). Every control-versus-marked comparison survives. $^\dagger$OpenBioLLM cannot reliably produce the structured or rewritten intermediate and is excluded from mitigation averages.}
\label{tab:app_mcnemar}
\end{table*}

\begin{table*}[t]
\centering
\fontsize{9.5}{12.5}\selectfont
\setlength{\tabcolsep}{9pt}
\resizebox{\ifdim\width>\textwidth\textwidth\else\width\fi}{!}{%
\begin{tabular}{lccccccc}
\toprule
\cellcolor{blue!10} & \multicolumn{4}{c}{\hc{Schema JSON parse rate}} & \hc{Questions with} & \multicolumn{2}{c}{\hc{Accuracy on those questions}} \\
\cmidrule(lr){2-5}\cmidrule(lr){7-8}
\hc{Model} & \hc{all} & \cellcolor{blue!10}$\boldsymbol{P_\alpha}$ & \cellcolor{blue!10}$\boldsymbol{P_\beta}$ & \cellcolor{blue!10}$\boldsymbol{P_\gamma}$ & \hc{all 3 parsed} & \hc{Schema} & \hc{Rewrite} \\
\midrule
Llama-3.2-3B-Instruct & 94.0 & 96.0 & 94.0 & 92.1 & 878 & .523 & .615 \\
medgemma-4b-it & 90.8 & 94.0 & 90.3 & 88.2 & 829 & .534 & .588 \\
Phi-4-mini-instruct & 97.8 & 98.9 & 98.1 & 96.4 & 952 & .473 & .515 \\
Qwen2.5-7B-Instruct & 96.8 & 98.1 & 95.8 & 96.6 & 943 & .559 & .588 \\
Mistral-7B-Instruct-v0.3 & 90.0 & 89.9 & 90.5 & 89.6 & 867 & .443 & .509 \\
Qwen3-8B & 97.9 & 98.6 & 97.7 & 97.3 & 967 & .569 & .639 \\
Ministral-8B-Instruct-2410 & 99.3 & 99.0 & 99.4 & 99.4 & 986 & .373 & .508 \\
Llama-3.1-8B-Instruct & 96.8 & 96.2 & 97.4 & 96.7 & 934 & .578 & .677 \\
Llama3-OpenBioLLM-8B$^\dagger$ & 84.2 & 79.9 & 86.2 & 86.5 & 687 & .206 & .099 \\
gemma-3-12b-it & 95.9 & 95.3 & 96.8 & 95.6 & 939 & .566 & .611 \\
Qwen2.5-14B-Instruct & 98.6 & 98.5 & 98.6 & 98.6 & 976 & .601 & .662 \\
Mistral-Small-24B-Instruct-2501 & 97.0 & 97.2 & 97.3 & 96.4 & 954 & .642 & .701 \\
\bottomrule
\end{tabular}%
}
\caption{Schema extraction. When the JSON does not parse, the answering call receives the raw extraction text. Restricted to questions where all three extractions parse, the rewrite remains more accurate in 11 of 11 models (mean +6.8 points), so the schema's accuracy loss is not a parse-failure artefact. $^\dagger$OpenBioLLM cannot reliably produce the structured or rewritten intermediate and is excluded from mitigation averages.}
\label{tab:app_parse}
\end{table*}

\end{document}